\documentclass[sigconf]{acmart}

\AtBeginDocument{%
  }

\copyrightyear{2026}
\acmYear{2026}
\setcopyright{cc}
\setcctype{by}
\acmConference[KDD '26]{Proceedings of the 32nd ACM SIGKDD Conference on Knowledge Discovery and Data Mining V.2}{August 09--13, 2026}{Jeju Island, Republic of Korea}
\acmBooktitle{Proceedings of the 32nd ACM SIGKDD Conference on Knowledge Discovery and Data Mining V.2 (KDD '26), August 09--13, 2026, Jeju Island, Republic of Korea}
\acmDOI{10.1145/3770855.3817488}
\acmISBN{979-8-4007-2259-2/2026/08}

\usepackage[utf8]{inputenc}
\usepackage{url}
\usepackage{xspace}
\usepackage{enumitem}
\usepackage[table,xcdraw]{xcolor}
\newcommand{\hide}[1]{}
\usepackage{multirow}
\usepackage{amsmath}
\usepackage{algorithm}
\usepackage{algorithmic}
\usepackage[normalem]{ulem}
\useunder{\uline}{\ul}{}
\usepackage{mathrsfs}
\usepackage{subfigure}
\usepackage{color}
\usepackage{tabularx}
\usepackage{bbding} 
\usepackage{array}
\usepackage{graphicx}
\usepackage{float}
\usepackage{makecell}
\usepackage{booktabs}
\usepackage{subcaption}
\usepackage{lipsum}
\usepackage{placeins}
\usepackage{tcolorbox}
\newcommand{\acronym}[1]{\underline{\textbf{#1}}}
\newcommand{\data}{\textbf{MAC}\xspace}
\newcommand{\presys}{\textbf{NATAL}\xspace}
\newcommand{\oursys}{\textbf{MoAE}\xspace}
\newcommand{\ourlib}{\textbf{PyMAL}\xspace}
\definecolor{demphcolor}{RGB}{144, 144, 144}
\definecolor{mygray}{gray}{0.4}
\definecolor{lightgray}{rgb}{0.9, 0.9, 0.9}

\usepackage[dvipsnames]{xcolor}

\makeatletter
\def\blfootnote{\gdef\@thefnmark{}\@footnotetext}
\makeatother

\title{MAC: A Conversion Rate Prediction Benchmark Featuring Labels Under Multiple Attribution Mechanisms}

\author{Jinqi Wu}
\email{jinqiwu001@gmail.com}
\authornote{Equal contribution.}
\affiliation{
  \institution{State Key Laboratory for Novel Software Technology, Nanjing University}
  \city{Nanjing}
  \country{China}
}
\affiliation{
  \institution{School of Intelligence Science and Technology, Nanjing University}
  \city{Suzhou}
  \country{China}
}

\author{Sishuo Chen}
\authornotemark[1]
\email{chensishuo.css@alibaba-inc.com}
\author{Zhangming Chan}
\email{zhangming.czm@alibaba-inc.com}
\author{Yong Bai}
\email{baiyong.by@alibaba-inc.com}
\affiliation{
  \institution{Taobao \& Tmall Group of Alibaba}
  \city{Beijing}
  \country{China}
}

\author{Lei Zhang}
\email{zl165646@alibaba-inc.com}
\author{Sheng Chen}
\email{chensheng.cs@alibaba-inc.com}
\author{Chenghuan Hou}
\email{jinyao@alibaba-inc.com}
\affiliation{
  \institution{Taobao \& Tmall Group of Alibaba}
  \city{Beijing}
  \country{China}
}

\author{Xiang-Rong Sheng}
\email{xiangrong.sxr@alibaba-inc.com}
\author{Han Zhu}
\email{zhuhan.zh@alibaba-inc.com}
\author{Jian Xu}
\email{xiyu.xj@alibaba-inc.com}
\affiliation{
  \institution{Taobao \& Tmall Group of Alibaba}
  \city{Beijing}
  \country{China}
}

\author{Bo Zheng}
\authornote{Corresponding authors.}
\email{bozheng@alibaba-inc.com}
\affiliation{
  \institution{Taobao \& Tmall Group of Alibaba}
  \city{Beijing}
  \country{China}
}

\author{Chaoyou Fu}
\authornotemark[2]
\email{bradyfu24@gmail.com}
\affiliation{
  \institution{State Key Laboratory for Novel Software Technology, Nanjing University}
  \city{Nanjing}
  \country{China}
}
\affiliation{
  \institution{School of Intelligence Science and Technology, Nanjing University}
  \city{Suzhou}
  \country{China}
}

\renewcommand{\shortauthors}{Jinqi Wu et al.}

\begin{document}

\begin{abstract}
Multi-attribution learning (MAL), which enhances model performance by learning from conversion labels yielded by multiple attribution mechanisms, has emerged as a promising learning paradigm for conversion rate (CVR) prediction.
However, the conversion labels in public CVR datasets are generated by a single attribution mechanism, hindering the development of MAL approaches.

To address this data gap, we establish the \acronym{M}ulti-\acronym{A}ttribution Ben\acronym{C}hmark (\data), the first public CVR dataset featuring labels from multiple attribution mechanisms.
Besides, to promote reproducible research on MAL, we develop \ourlib, an open-source library covering a wide array of baseline methods.
We conduct comprehensive experimental analyses on \data and reveal three key insights:
\textbf{(1)} MAL brings consistent performance gains across different attribution settings, especially for users featuring long conversion paths.
\textbf{(2)} The performance growth scales up with objective complexity in most settings; however, when predicting first-click conversion targets, simply adding auxiliary objectives is counterproductive, underscoring the necessity of careful selection of auxiliary objectives.
\textbf{(3)} Two architectural design principles are paramount: first, to fully learn the multi-attribution knowledge, and second, to fully leverage this knowledge to serve the main task.
Motivated by these findings, we propose \acronym{M}ixture \acronym{o}f \acronym{A}symmetric \acronym{E}xperts (\oursys), an effective MAL approach incorporating multi-attribution knowledge learning and main task-centric knowledge utilization.
Experiments on \data show that \oursys substantially surpasses the existing state-of-the-art MAL method.
We believe that our benchmark and insights will foster future research in the MAL field.
Our \data benchmark and the \ourlib algorithm library are publicly available at \color{blue}{\url{https://github.com/alimama-tech/PyMAL}}.

\end{abstract}

\ccsdesc[500]{Applied computing~Electronic commerce}
\ccsdesc[500]{Information systems~Computational advertising}
\ccsdesc[500]{Computing methodologies~Ranking}

\keywords{Multi-Attribution Learning, Conversion Rate Prediction, Online Advertising, Dataset and Benchmark}

\maketitle

\section{Introduction}

\begin{figure}[!t]
\centering 
\includegraphics[width=\columnwidth]{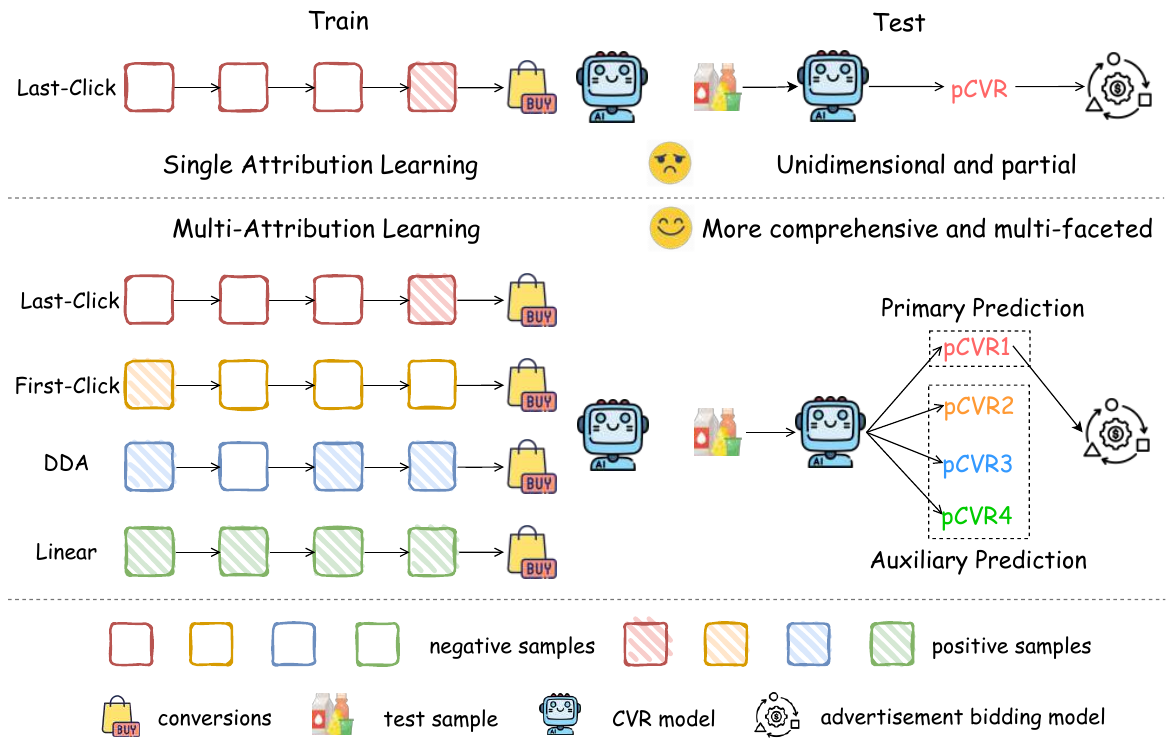}
\caption{Illustration of single-attribution learning and multi-attribution learning (MAL) for CVR prediction. Our work paves the way for studying MAL by providing the first public benchmark, systematic evaluation, and valuable insights. }
\Description{A conceptual illustration comparing single-attribution learning and multi-attribution learning for conversion rate prediction. The figure shows how different attribution mechanisms provide different supervision signals, motivating the proposed benchmark for studying multi-attribution learning.}
\label{fig:mac_intro} 
\vspace{-0.4cm}
\end{figure} 

Conversion rate (CVR) prediction is a cornerstone of online advertising systems, directly influencing bidding efficiency, traffic allocation, and ultimately, platform revenue~\cite{chapelle2014modeling,lu2017practical}.
At its core, the logic of label generation and model optimization is determined by \textbf{attribution mechanisms}~\cite{shao2011data,yao2022causalmta}--the protocols allocating conversion credits across user touchpoints (usually operationalized as ad clicks).

Currently, nearly all existing CVR prediction models, in both industry and academia, are trained and evaluated under a \textbf{single-attribution assumption}~\cite{chen2025see}, where conversion labels are derived from a single attribution mechanism.
This paradigm constrains models to learning a partial view of touchpoint value, failing to capture the user's complex conversion mindset holistically.

Recently, \textbf{multi-attribution learning (MAL)} ~\cite{chen2025see} has emerged as a promising new paradigm that enhances model performance by jointly learning from conversion labels generated by multiple attribution mechanisms, such as first-click, last-click, and linear attribution rules, and data-driven attribution (DDA) models~\cite{yao2022causalmta,lewis2025amazon}. 
By integrating multi-faceted auxiliary signals from diverse attribution views, MAL enables the model to develop a more comprehensive and holistic understanding of touchpoint contributions, which leads to significantly higher prediction accuracy under the system-optimized attribution mechanism and ultimately drives substantial growth in user satisfaction and platform revenue.
Figure~\ref{fig:mac_intro} compares the conventional single-attribution learning and the novel multi-attribution learning paradigm.

However, despite its considerable promise, the advancement of research on MAL is constrained by a critical data bottleneck: to the best of our knowledge, all publicly available CVR datasets, such as Criteo~\cite{chapelle2014modeling}, Ali-CCP~\cite{esmm}, and Taobao~\cite{zhu2018learning}, only provide conversion labels from a single attribution mechanism.

\begin{figure}[!t]
\centering 
\includegraphics[width=\columnwidth]{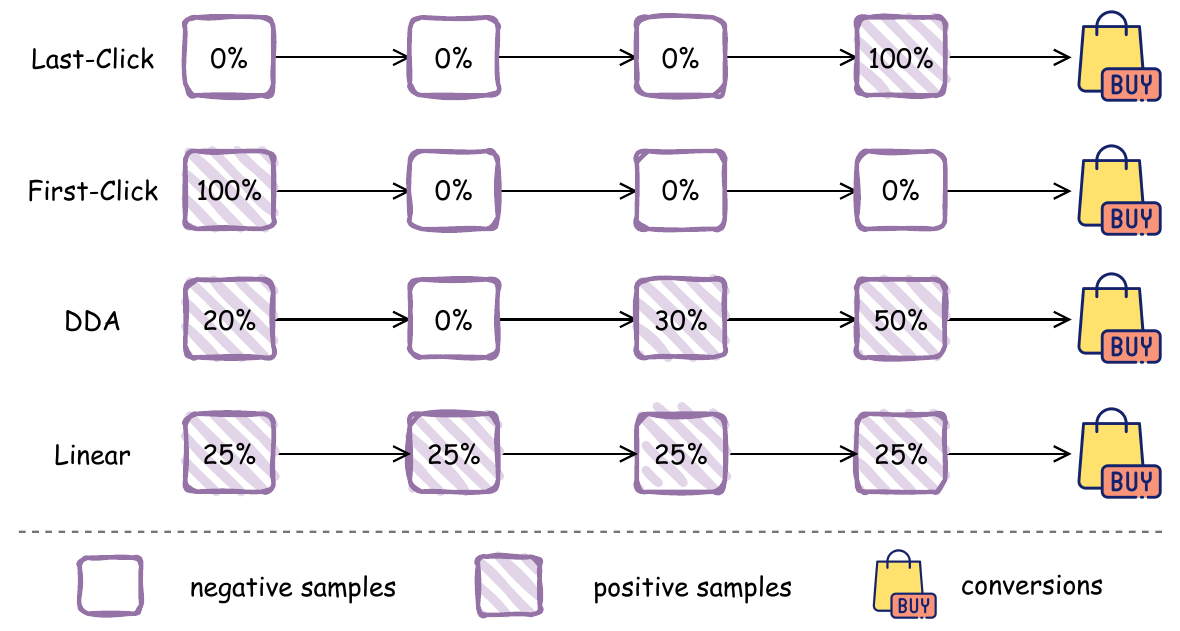}
\caption{The four conversion attribution mechanisms covered in our \data benchmark.} 
\Description{An illustrative example of four conversion attribution mechanisms included in the benchmark, showing how conversion credit is assigned differently across user interaction paths.}
\label{fig:mechanism_example} 
\end{figure}

To address this, we establish the \acronym{M}ulti-\acronym{A}ttribution Ben\acronym{C}hmark (\data), the first CVR prediction benchmark \textbf{featuring conversion labels from multiple attribution mechanisms} to our knowledge. 
For each ad click, \data provides conversion labels yielded by four typical attribution mechanisms: \textit{last-click, first-click, linear, and data-driven attribution (DDA)}, which are illustrated in Figure~\ref{fig:mechanism_example}.
\data is collected from the advertising system of Taobao, Alibaba, and rigorously
anonymized for public release, providing a reliable testbed for assessing MAL approaches for CVR prediction.

Moreover, to facilitate reproducible research on MAL, we release \ourlib, an open-source library covering representative baselines, ranging from classic multi-task models~\cite{mmoe,ple,wang2024home} to the cutting-edge NATAL model~\cite{chen2025see}\footnote{NATAL denotes the k\acronym{N}owledge \acronym{A}ggregation with the car\acronym{T}esian \acronym{A}uxi\acronym{L}iary training model proposed in ~\citet{chen2025see}.}.
Using \textbf{PyMAL}, we perform a systematic evaluation on \data and discover three key insights:
\begin{enumerate}[leftmargin=*]
    \item \textbf{The Generality of MAL}:  MAL consistently improves prediction accuracy across different target attribution mechanisms, with the most notable gains observed in users featuring long conversion journeys.
    \item \textbf{The Effect of Auxiliary Objective Choice in MAL}: While stacking more auxiliary targets from different attribution mechanisms typically enhances performance, it is not always the case. 
    Under the first-click setting, indiscriminate addition of auxiliary tasks hurts model generalization, highlighting the importance of strategically selecting complementary attribution views.~\looseness=-1
    \item \textbf{Structure Design Principles for Effective MAL}: We identify two key principles: (i) \textbf{fully learning multi-attribution knowledge}, as achieved by the mixture-of-experts (MoE) structures in MTL models like MMoE~\cite{mmoe} and PLE~\cite{ple}; (ii) \textbf{leveraging this knowledge in a main-task-prioritized manner}, as realized through asymmetric knowledge transfer in NATAL~\cite{chen2025see}.
\end{enumerate}

Inspired by the identified principles, we propose \textbf{M}ixture \textbf{o}f \textbf{A}symmetric \textbf{E}xperts (\oursys), a novel MAL model that simultaneously satisfies both design principles by incorporating an MoE backbone on the bottom for comprehensive multi-attribution knowledge learning, complemented by a main-task-prioritized asymmetric transfer module on the top.
Extensive experiments on \data show that \oursys significantly outperforms existing MAL models across all settings, raising the GAUC metric by up to \textbf{0.39pt}\footnote{``pt'' is short for percentage points throughout this paper.}.
Besides improving the main task metric, \oursys also surpasses the existing state-of-the-art model NATAL~\cite{chen2025see} in terms of auxiliary task metrics.
This confirms that our MoAE method more effectively captures the incremental knowledge from multi-attribution labels.

We summarize our contributions as follows:
\begin{itemize}[leftmargin=10pt,topsep=2pt]
\item \textbf{Data Resource}: We release \textbf{the first multi-attribution CVR benchmark} named \data.
Covering conversion labels derived from four representative attribution mechanisms, \data fills the key data gap in multi-attribution learning and provides a foundational benchmark for future research.

\item \textbf{Software Toolkits and Comprehensive Benchmarking}: We release \textbf{PyMAL}, an open-source library covering typical MAL approaches, and benchmark them on \data. 
Our thorough analysis of the benchmarking results confirms the universality of MAL and offers actionable insights for designing critical components in MAL, such as auxiliary objectives and model architecture.

\item \textbf{Effective Approach}: We propose \oursys, a novel MAL model that integrates the MoE architecture for efficiently capturing multi-attribution knowledge and a main-task-oriented knowledge transfer mechanism.
Extensive evaluation results show that \oursys substantially outperforms existing methods on \data.~\looseness=-1
\end{itemize}

\section{Background and Related Works}

\subsection{CVR Prediction and Attribution Mechanism}
Conversion rate (CVR) prediction~\cite{chapelle2014modeling,lu2017practical,esmm,chan2023capturing} plays a vital role in online advertising.
Researchers have pursued numerous avenues to improve CVR prediction models, such as joint learning with other tasks~\cite{esmm,wang2022escm2,zhao2023ecad,luo2026modeling}, model structure~\cite{zhuang2025practice,zhu2025rankmixer}, and delayed feedback modeling~\cite{yang2021capturing,defer,defuse,liu2024online}.
Although \textbf{attribution mechanisms}~\cite{shao2011data,yao2022causalmta}, namely the rules for allocating conversion credits to ad clicks, directly determine the learning goals of CVR prediction models, they are overlooked in existing studies, which all rely on a single attribution mechanism.
\citet{chen2025see} proposes multi-attribution learning (MAL) and achieves significant performance gains on in-house data, highlighting the potential of the MAL paradigm.
Nevertheless, existing public CVR datasets~\cite{chapelle2014modeling,esmm,zhu2018learning} all use a single attribution mechanism to generate labels, hindering advances in MAL.
To the best of our knowledge, we release the first CVR dataset with multi-attribution labels, and further contribute the first algorithmic codebase and in-depth analysis for MAL approaches.

\subsection{Multi-Task Learning for Recommendation}
Multi-task learning (MTL)~\cite{zhang2021survey,caruana1997multitask,misra2016cross,star} seeks to train a single model for predicting multiple targets simultaneously.
Researchers have made extensive efforts to apply MTL to recommendation tasks, ranging from general structure design~\cite{mmoe,ple,li2023adatt,wang2024home,yu2021content} to specific task grouping~\cite{esmm,zhao2023ecad,wang2022escm2,chen2025see,li2026delayed,chan2020selection}.
\citet{chen2025see} regard MAL as an MTL problem and apply classical MTL models to MAL, such as MMoE~\cite{mmoe} and PLE~\cite{ple}, where they underperform the NATAL model tailored for MAL.
In this study, we will delve deeper into the strengths and weaknesses of the MTL methods within the MAL context, aiming to collect insights for future architectural design.

\section{The \data Benchmark}

\subsection{The MAL Problem Setup}
In CVR prediction, the conventional paradigm assumes a \textbf{single attribution mechanism} $\mathcal{A}$ (e.g., last-click or linear) that assigns a binary label $y^\mathcal{A} \in \{0,1\}$ to each ad click, indicating whether the click is credited for a subsequent conversion within a fixed attribution window. However, this view is inherently limited: different attribution mechanisms encode distinct perspectives on user intent and touchpoint contribution, and no single rule fully captures the true conversion causality.
To overcome this limitation, we formalize \textbf{multi-attribution learning (MAL)} as follows. Let $\mathcal{M} = \{\mathcal{A}_1, \mathcal{A}_2, \dots, \mathcal{A}_K\}$ be a set of $K$ attribution mechanisms (e.g., last-click, first-click, linear, and data-driven attribution (DDA)).
For each click instance $\mathbf{x}$, there is a set of \textbf{continuous attribution weights} $\{w^{\mathcal{A}_1}, w^{\mathcal{A}_2}, \dots, w^{\mathcal{A}_K}\}$, where $w^{\mathcal{A}_i} \in \mathbb{R}_{\geq 0}$ denotes the contribution of the click to related conversions under the mechanism $\mathcal{A}_i$, computed over a fixed attribution window.

In MAL, one attribution mechanism $\mathcal{A}_t \in \mathcal{M}$ is designated as the \textbf{target attribution mechanism} according to business requirements.
Predicting the conversion labels under $\mathcal{A}_t$  is the \textbf{primary task}, while the remaining $K-1$ mechanisms serve as \textbf{auxiliary attribution mechanisms}, and the labels yielded by them are \textbf{auxiliary targets}. The goal is to train a prediction model $f_\theta(\cdot)$ such that the estimate $\hat{y}^{\mathcal{A}_t} = f_\theta(\mathbf{x})$ accurately predicts the binary conversion label derived from the attribution weight $w^{\mathcal{A}_t}$ by jointly leveraging supervision signals from all $K$ attribution views. This setup enables the model to acquire a more holistic understanding of post-click conversion behavior.

\subsection{Benchmark Construction}
\subsubsection{Overview} \label{subsubsec:data_overview}
Existing public CVR datasets~\cite{chapelle2014modeling,esmm,zhu2018learning} only provide labels under one fixed attribution mechanism, making systematic research on MAL infeasible. 
To break the data bottleneck, we introduce the \acronym{M}ulti-\acronym{A}ttribution Ben\acronym{C}hmark (\data), \textbf{the first open-source CVR dataset that provides multi-dimensional attribution weights under different attribution mechanisms}.
\data is sampled from real-world industrial logs of the advertising system of  Taobao and covers four types of popular attribution mechanisms: last-click, first-click, linear, and data-driven attribution (DDA)~\cite{yao2022causalmta}.
This characteristic enables a systematic investigation of MAL approaches for the research community.
We summarize the statistics and key features of \data and existing benchmarks in Table~\ref{tab:data_stat} and list the positive sample ratios under different mechanisms in \data in Table~\ref{tab:pos_ratio}.
The rules of conversion attribution in each attribution mechanisms are given as follows:
\begin{itemize}[leftmargin=*]
    \item \textbf{Last-Click}: Concentrates 100\% attribution weight on the final click preceding conversion.
    \item \textbf{First-Click}: Attributes full credit  to the initial click.
    \item \textbf{Linear}: Distributes weights uniformly across all touchpoints. 
    \item \textbf{Data-driven attribution (DDA)}: Learns attribution weight allocation through causal inference models~\cite{zhou2019deep,yao2022causalmta,kumar2020camta,bencina2025lidda}.
\end{itemize} 


\begin{table}[t]
\centering
\caption{Summary of existing CVR prediction datasets and our \data benchmark. (Notes: BS = Behavior Sequence, MF = Multimodal Feature, MA = Multi-Attribution Labels.)}
\label{tab:data_stat}
\label{tab:main_statistics}
\resizebox{0.45\textwidth}{!}{
\begin{tabular}{@{}l|ccccccc@{}}
\toprule
\multirow{2}{*}{\textbf{Dataset}} &
\multirow{2}{*}{\textbf{\#Clicks}} & 
\multirow{2}{*}{\textbf{\#Items}} & 
\multirow{2}{*}{\textbf{\#Users}} & 
\multirow{2}{*}{{\textbf{BS}}} & 
\multirow{2}{*}{{\textbf{MF}}} &
\multirow{2}{*}{{\textbf{MA}}} \\
& & & & & & \\
\midrule
Criteo~\cite{chapelle2014modeling} & 3.6M & 15.9M & 5.0K & \XSolidBrush & \XSolidBrush & \XSolidBrush\\
Ali-CCP~\cite{esmm} & 3.4M & 4.3M & 0.4M & \XSolidBrush & \XSolidBrush & \XSolidBrush\\
Taobao~\cite{zhu2018learning} & 100M & 4.0M & 1.0M & \XSolidBrush & \XSolidBrush & \XSolidBrush\\
\midrule
\data & 79M & 15.1M & 0.8M & \Checkmark & \Checkmark & \Checkmark \\
\bottomrule
\end{tabular}}
\end{table}

\begin{table}[t]
\centering
\caption{The positive sample ratio under different attribution mechanisms in our \data benchmark.}
\label{tab:pos_ratio}
\resizebox{0.45\textwidth}{!}{
\begin{tabular}{@{}l|cccc@{}}
\toprule
\textbf{} &
\textbf{Last-Click} &
\textbf{First-Click} & 
\textbf{DDA} & 
\textbf{Linear} \\
\midrule
\textbf{Positive Ratio}  & 1.6\% & 1.8\% & 4.3\% & 5.3\% \\
\bottomrule
\end{tabular}}
\end{table}

\subsubsection{Data Sampling Criteria}
\data covers 21 consecutive days of traffic on Taobao, one of the world's largest e-commerce platforms.
To ensure that the dataset size is suitable for academic research, we performed stratified sampling, where we assigned higher
sampling rates to highly active users and lower rates to less active ones.
The resulting \data dataset includes 0.8 million users, 79 million clicks, and 15.1 million distinct items.
As shown in Table~\ref{tab:data_stat}, \data is comparable to or larger than existing datasets~\cite{chapelle2014modeling,esmm,zhu2018learning}.


\subsubsection{Feature and Label Schema}
Each  ad click sample in \data contains three groups of signals: 
\begin{itemize}
[leftmargin=10pt,topsep=2pt]
    \item \textit{Categorical features}: 7 user features including user ID and profile features, 10 item features such as item, shop, category, and adgroup IDs, and 3 context features such as the ad scenario ID.
    \item \textit{Behavior sequence features}: the user purchase item sequence with a maximum truncated length of 20.
    For each item, we provide the item, shop, and category IDs along with its visual similarity score to the target advertised item, using the features from an image encoder pre-trained through contrastive learning~\cite{sheng2024enhancing}.
    \item \textit{Multi-attribution conversion labels}: \data provides continuous attribution weights for each ad click under all four attribution mechanisms as stated in \S~\ref{subsubsec:data_overview}, derived from the conversion paths within a fixed, confidential attribution window.
    We use the CausalMTA model~\cite{yao2022causalmta} as the DDA model and use direct conversion attribution, which means attributing conversions only to clicks on ads within the same item.
\end{itemize}

\subsubsection{Data Compliance and Availability}
Our \data benchmark has undergone rigorous anonymization to remove personally sensitive information.
All personally identifiable information and business details, such as user, shop, and adgroup identifiers, are irreversibly hashed.
The dataset is constructed from historical traffic logs and does not contain real-time operational metrics or business conditions due to the user sampling procedure.
We release \data as a public benchmark to support future research on CVR prediction on \textcolor{blue}{\href{https://huggingface.co/datasets/alimamaTech/MAC}{https://huggingface.co/datasets/alimamaTech/MAC}.}

\subsection{Evaluation Metrics}

Following industrial practice \cite{chen2025see}, we regard CVR prediction as a
binary classification problem derived from attribution weights:
\begin{itemize}
    \item \textbf{Positive Samples.} Samples with positive attribution weights are treated as positive samples.
    \item \textbf{Negative Samples.} Samples with zero attribution weights are treated as negative samples.
\end{itemize}

For evaluating model performance, we report the widely-adopted \textbf{AUC} and Group AUC (\textbf{GAUC}) metrics~\cite{schutze2008introduction,din,ShengGCYHDJXZ2023JRC,BianWRPZXSZCMLX2022CAN} of the primary task.
AUC is a widely used ranking metric that evaluates a model’s ability to distinguish positive and negative samples, making it well-suited for CVR prediction.
The definition of AUC is given in Eq.~\eqref{eqn:auc}
\begin{equation}
\label{eqn:auc} 
\begin{aligned} 
    \text{AUC} = \frac{1}{n_+ n_-} \sum_{i=1}^{n_+} \sum_{j=1}^{n_-} \left[ \mathbb{I}(y_i^+ > y_j^-) + \frac{1}{2} \mathbb{I}(y_i^+ = y_j^-) \right]
\end{aligned}
\end{equation}
Here $n_+$ and $n_-$ denote the numbers of positive and negative samples, respectively, $y_i^+$ and $y_j^-$ stand for the predicted conversion probabilities for the $i$-th positive and $j$-th negative sample, respectively, and $\mathbb{I}(\cdot)$ is the indicator function.

Moreover, we report GAUC, which captures the model’s ability to rank ad clicks within individual user groups and serves as the primary evaluation criterion in our production system. 
It has been empirically shown to align more closely with online performance than AUC.
The definition of GAUC is given in Eq.~\eqref{eqn:gauc}:
\begin{equation}
\label{eqn:gauc} 
\begin{aligned} 
    \textrm{GAUC} = \frac{\sum_{u=1}^U \# \textrm{click}(u) \times \textrm{AUC}_u}{\sum_{u=1}^U \# \textrm{click}(u)}
\end{aligned}
\end{equation}
Here $U$ represents the number of users, $\#\textrm{click}(u)$ denotes the number of clicks for the $u$-th user, and $\textrm{AUC}_u$ is the AUC computed on the samples from the $u$-th user.

\section{The PyMAL Library}

To foster reproducible and comparable research on multi-attribution learning (MAL) for CVR prediction, we establish \textbf{PyMAL}, an open-source Python library built on top of PyTorch~\cite{paszke2019pytorch} available at \textcolor{blue}{\href{https://github.com/alimama-tech/PyMAL}{GitHub}}.
It provides a unified, modular, and extensible framework that implements a comprehensive suite of MAL baselines, and it can be easily extended to incorporate ongoing efforts in the MAL field. 
The library supports rapid prototyping and fair comparison of MAL approaches, which paves the way for studies on MAL in the community.

\begin{figure}[t]
\centering 
\includegraphics[width=0.85\columnwidth]{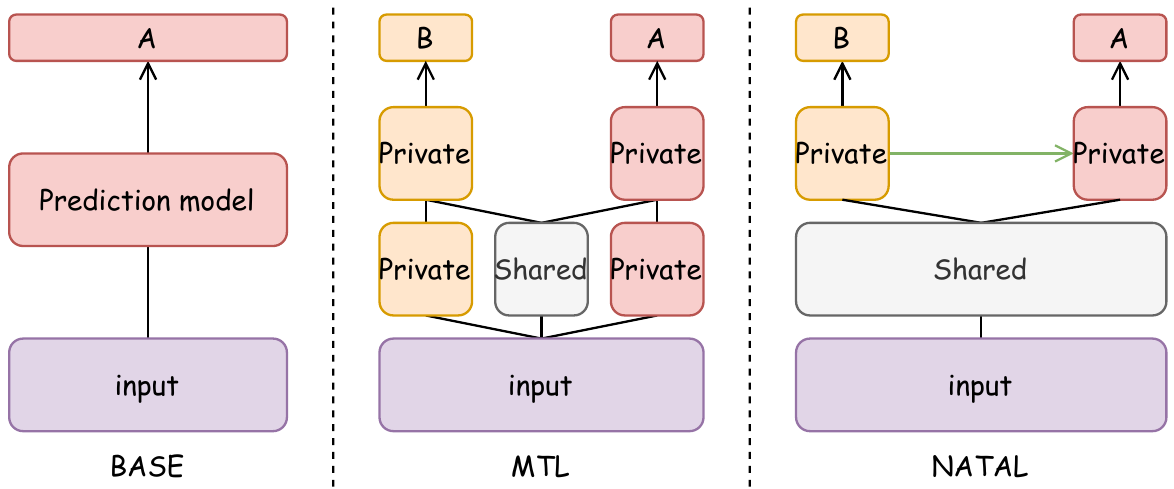}
\caption{Illustration of three families of baselines. A is the primary target under the target attribution mechanism, and B is an auxiliary target from other attribution mechanisms.}
\Description{A schematic comparison of three baseline model families for multi-attribution learning. The figure uses A to denote the primary prediction target under the target attribution mechanism and B to denote an auxiliary target from other attribution mechanisms.}
\label{fig:model_compare} 
\end{figure}

We categorize the implemented models in \textbf{PyMAL} into three representative families as follows, reflecting the evolution of modeling paradigms in MAL.
Figure~\ref{fig:model_compare} illustrates their model structures.

\textbf{First, the single-attribution learning baseline.}
The \textbf{BASE} model is trained solely on the labels generated by the target attribution mechanism (e.g., last-click), ignoring the auxiliary signals from other attribution mechanisms.
Regarding model architecture, it comprises target attention~\cite{din} for sequence modeling, SimTier for multimodal modeling~\cite{sheng2024enhancing}, and an MLP predictor for outputting the predicted CVR value, following prior industrial practice~\cite{chen2025see}.

\textbf{Second, the multi-task learning (MTL) models adapted for MAL.}
These models jointly learn the primary task together with one or more auxiliary attribution tasks, using shared or structured representations to capture cross-task knowledge:
\begin{itemize}[leftmargin=10pt,topsep=2pt]
    \item \textbf{Shared-Bottom}~\cite{caruana1997multitask}: A classic architecture featuring a shared bottom embedding layer followed by task-specific output towers.
    \item \textbf{MMoE}~\cite{mmoe}: A mixture-of-experts model that uses task-specific gating networks to dynamically combine shared expert subnetworks, enabling flexible knowledge sharing.
    \item \textbf{PLE}~\cite{ple}: A progressive layered extraction architecture that explicitly separates shared and task-specific experts to reduce negative transfer and improve task cooperation.
    \item \textbf{HoME}~\cite{wang2024home}: An enhanced MoE variant that introduces Self-Gate and Feature-Gate mechanisms to adaptively modulate expert selection based on input feature granularity, coupled with Swish activation for better nonlinearity modeling.
\end{itemize}

\textbf{Last, the state-of-the-art NATAL model tailored for MAL.}
NATAL~\cite{chen2025see}, short for k\acronym{N}owledge \acronym{A}ggregation with the car\acronym{T}esian \acronym{A}uxi\acronym{L}iary training, goes beyond conventional MTL by explicitly modeling attribution interactions and prioritizing the primary objective.
Concretely, it incorporates three key components: (1) \textit{Attribution Knowledge Aggregation} (AKA) for fusing supervision from multiple attribution mechanisms via intermediate feature transfer; (2) \textit{Primary Target Predictor} (PTP) for generating the prediction under the target attribution mechanism; and (3) \textit{Cartesian-based Auxiliary Training} (CAT), a Cartesian product-based auxiliary task that integrates higher-order attribution interactions, whose details are given in Appendix~\ref{app:cat}.

\section{Method: Mixture of Asymmetric Experts} \label{sec:method}

\subsection{Motivation: Structure Design Principles} \label{subsec:moae_motivation}

To understand the pros and cons of prior methods, we record their performance under both the target and auxiliary attribution mechanisms.
As shown in Table~\ref{tab:motivation_moae}, the advanced MTL model equipped with the MoE structure, e.g., PLE and HoME, not only raises the primary task performance, but also improves the auxiliary task metrics compared with Shared-Bottom.
This is reasonable, as it is known that MoE can capture task commonalities and reduce gradient conflicts, thereby improving the performance of all subtasks in MTL~\cite{mmoe,ple,wang2024home}.
In contrast, NATAL discards the MoE structure and introduces asymmetric knowledge transfer from auxiliary towers to the main task tower for enhancing main task-prioritized knowledge utilization~\cite{chen2025see}.
As a result, although NATAL gives the best main task performance among baselines, it underperforms PLE and HoME in terms of auxiliary task performance.
Therefore, we contend that a successful MAL model should meet two principles:
\begin{tcolorbox}[colback=cyan!5!white, colframe=cyan!45!blue!60, title=\textbf{Structure Design Principles}]
\begin{enumerate}[label=\arabic*.,left=-2pt] 
\item Fully learn multi-attribution knowledge underlying multiple attribution mechanisms, as done in PLE and HoME;~\looseness=-1
\item Fully leverage this knowledge in a main-task-prioritized manner, as done in NATAL.
\end{enumerate}
\end{tcolorbox}

\subsection{Our MoAE Model}

\begin{figure}[tbp]
\centering 
\includegraphics[width=\columnwidth]{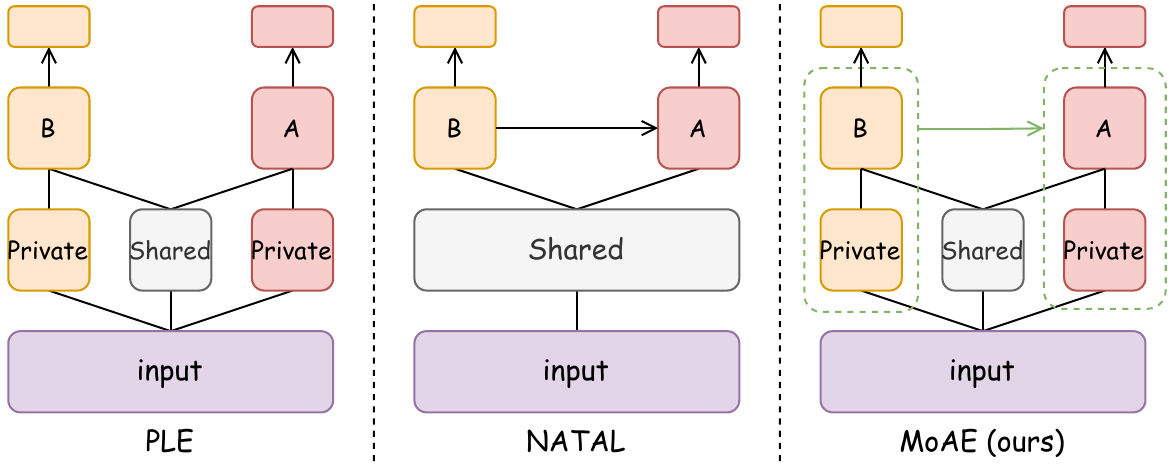}
\caption{Information flow comparison between cutting-edge baselines (PLE and NATAL) and our \oursys model. A denotes the main target under the target attribution mechanism, and B is an auxiliary target from other attribution mechanisms.}
\Description{A diagram comparing the information flow of PLE, NATAL, and the proposed model. The figure shows how the main target A under the target attribution mechanism and the auxiliary target B from other attribution mechanisms are used differently by each model.}
\label{fig:model} 
\end{figure}

To satisfy the two aforementioned principles, we
devise \textbf{M}ixture \textbf{o}f \textbf{A}symmetric \textbf{E}xperts (\oursys), a new multi-attribution learning model integrating the strengths of the baselines analyzed in \S~\ref{subsec:moae_motivation}.
Concretely, as illustrated in Figure~\ref{fig:model}~\footnote{We only show one auxiliary attribution mechanism here for brevity. When there are multiple auxiliary attribution mechanisms, \oursys allocates one private expert and an MLP predictor for each auxiliary attribution mechanism.}, \oursys consists of three components:
\textbf{(1)} An MoE backbone that learns rich attribution-specific representations. 
Inspired by PLE~\cite{ple}, we introduce a shared expert for learning public conversion patterns and attribution-specific experts for learning unique patterns underlying different attribution mechanisms.
\textbf{(2)} A main–task–centric feature transfer module that aggregates auxiliary knowledge into the main task predictor in an asymmetric way.
\textbf{(3)} MLP predictors that produce predictions for each attribution mechanism.

\begin{table}[t]
\centering
\caption{The model performance in terms of GAUC when last-click serves as the target attribution mechanism. MoE denotes the mixture-of-experts structure, and KT stands for main-task-prioritized knowledge transfer. The best results are indicated in \textbf{bold} and the second best results are \underline{underlined}.}
\label{tab:motivation_moae}
\resizebox{0.475\textwidth}{!}{
\begin{tabular}{@{}l|cc|c|ccc@{}}
\toprule 
\multirow{2}{*}{\textbf{Model}}  & \multirow{2}{*}{\textbf{MoE}} & \multirow{2}{*}{\textbf{KT}}  & \textbf{Primary} & \multicolumn{3}{c}{\textbf{Auxiliary}} \\ 
 & & & Last & First & DDA & Linear \\
\midrule
Base       &  \XSolidBrush    &   \XSolidBrush      & 0.7424 & - & - & - \\
Shared-Bottom   &  \XSolidBrush  & \XSolidBrush  & 0.7571 & 0.6844 & 0.7579 & 0.7619 \\
MMoE  &  \Checkmark   &  \XSolidBrush            & 0.7571 & 0.6843 & 0.7586 & 0.7626 \\
PLE   &   \Checkmark  &   \XSolidBrush         & 0.7583 & \textbf{0.6900} & 0.7620 & 0.7660 \\
HoME    &   \Checkmark  & \XSolidBrush         & 0.7594 & 0.6871 & \underline{0.7624} & \underline{0.7666} \\
NATAL    & \XSolidBrush   &   \Checkmark      & \underline{0.7613} & 0.6842 & 0.7620 & 0.7650 \\ 
\textbf{MoAE (ours)}  &   \Checkmark  &   \Checkmark  & \textbf{0.7636} & \underline{0.6877} & \textbf{0.7637} & \textbf{0.7668} \\
\bottomrule
\end{tabular}}
\end{table}

As shown in Table~\ref{tab:motivation_moae}, our MoAE model not only reaches the best primary task performance, but also reaches superior performance in terms of the GAUC metrics under the auxiliary attribution mechanisms, which suggests that \oursys more fully fits the objectives under each attribution mechanism, aligning with its design motivation.
We further show the generality of \oursys in the following experiments.

\section{Experimental Results and Analysis}

\begin{table*}[t]
\centering
\caption{The CVR ranking performance under different target attribution mechanisms.}
\label{tab:main_results}
\resizebox{0.85\textwidth}{!}{
\begin{tabular}{@{}l|cc|cc|cc|cc@{}}
\toprule 
\textbf{Primary Target} & \multicolumn{2}{c}{\textbf{Last-Click}} & \multicolumn{2}{|c}{\textbf{First-Click}} & \multicolumn{2}{|c}{\textbf{DDA}} & \multicolumn{2}{|c}{\textbf{Linear}} \\
\midrule
\textbf{Method/Metrics} & \multicolumn{1}{c}{\textbf{AUC $\uparrow$}}   & \textbf{GAUC $\uparrow$}  & \multicolumn{1}{c}{\textbf{AUC $\uparrow$}}   & \textbf{GAUC $\uparrow$}  & \multicolumn{1}{c}{\textbf{AUC $\uparrow$}}   & \textbf{GAUC $\uparrow$}  & \multicolumn{1}{c}{\textbf{AUC $\uparrow$}}   & \textbf{GAUC $\uparrow$} \\
\midrule
BASE                & 0.8494 & 0.7424 & 0.8311 & 0.6969 & 0.8435 & 0.7470 & 0.8503 & 0.7602\\
Shared-Bottom ~\cite{caruana1997multitask} & 0.8628 & 0.7571 & 0.8303 & 0.6979 & 0.8514 & 0.7575 & 0.8515 & 0.7631 \\
MMoE ~\cite{mmoe} & 0.8642 & 0.7571 & 0.8314 & 0.6977 & 0.8507 & 0.7555 & 0.8512 & 0.7634 \\
PLE ~\cite{ple}    & 0.8645 & 0.7583 & 0.8316 & 0.6986 & 0.8509 & 0.7574 & 0.8527 & 0.7638 \\
HoME ~\cite{wang2024home}      & 0.8661 & 0.7594 & 0.8316 & 0.6985 & 0.8524 & 0.7589 & 0.8542 & \underline{0.7643} \\
NATAL ~\cite{chen2025see} & \underline{0.8671} & \underline{0.7613} & \underline{0.8319} & \underline{0.6990} & \underline{0.8546} & \underline{0.7617} & \underline{0.8548} & \underline{0.7643} \\
\midrule
\textbf{\oursys (ours)}   & \textbf{0.8694} & \textbf{0.7636}  & \textbf{0.8325} & \textbf{0.7003} & \textbf{0.8576} & \textbf{0.7644} & \textbf{0.8568} & \textbf{0.7682} \\
{$\Delta$ (vs BASE)}    & \textcolor{Green}{\textbf{+2.00pt}} & \textcolor{Green}{\textbf{+2.12pt}}  & \textcolor{Green}{\textbf{+0.14pt}} & \textcolor{Green}{\textbf{+0.34pt}} & \textcolor{Green}{\textbf{+1.41pt}} & \textcolor{Green}{\textbf{+1.74pt}} & \textcolor{Green}{\textbf{+0.65pt}} & \textcolor{Green}{\textbf{+0.80pt}} \\
{$\Delta$ (vs NATAL)}    & \textcolor{Green}{\textbf{+0.23pt}} & \textcolor{Green}{\textbf{+0.23pt}}  & \textcolor{Green}{\textbf{+0.06pt}} & \textcolor{Green}{\textbf{+0.13pt}} & \textcolor{Green}{\textbf{+0.30pt}} & \textcolor{Green}{\textbf{+0.27pt}} & \textcolor{Green}{\textbf{+0.20pt}} & \textcolor{Green}{\textbf{+0.39pt}} \\
\bottomrule
\end{tabular}}
\end{table*}

\begin{figure*}[thb] \centering
    \includegraphics[width=0.24\textwidth]{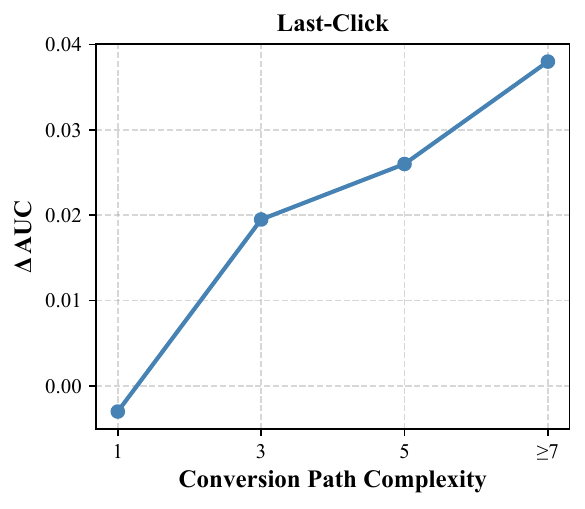}
    \includegraphics[width=0.24\textwidth]{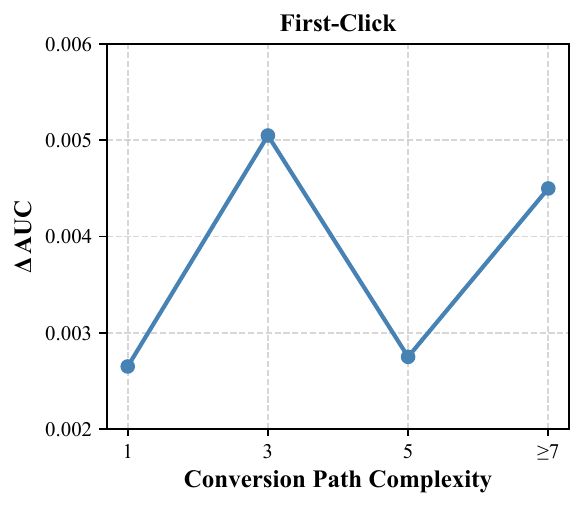}
    \includegraphics[width=0.24\textwidth]{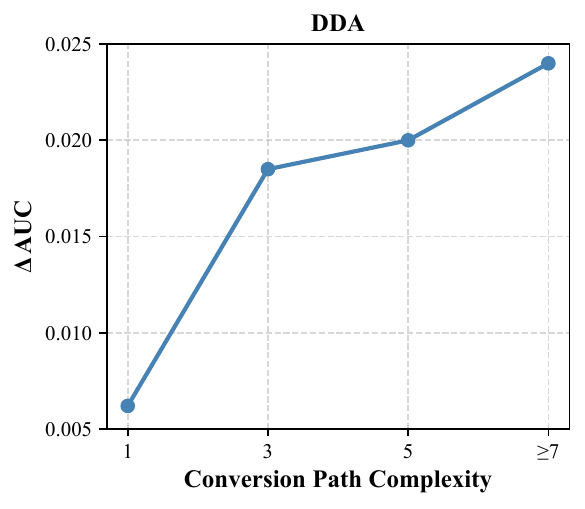}
    \includegraphics[width=0.24\textwidth]{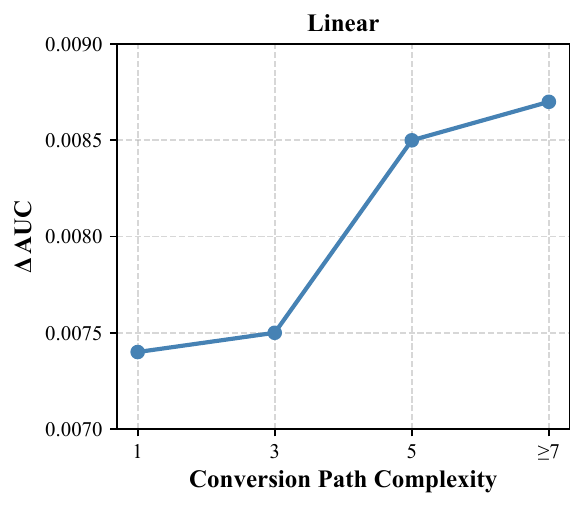}
    \caption{The AUC growth of users grouped by conversion path complexity.}
    \Description{Four line charts showing AUC growth for user groups with different conversion path complexity levels under Last-Click, First-Click, DDA, and Linear attribution mechanisms. The x-axis represents conversion path complexity, and the y-axis represents AUC growth.}
    \label{fig:user_groups}
\end{figure*}

\subsection{Experimental Setup}

\subsubsection{Training Hyperparameters}
We train all models using the Adam optimizer~\cite{Kingma2014AdamAM} for one epoch~\cite{Zhang2022TowardsUT} on the \data, where the last-day data serve as the test set and the preceding data are used for training. 
The batch size is 4096 for all models, as exploratory experiments showed that this value works well across different settings.
Regarding the learning rate, we search for the best value over the set $\text{lr} \in \{0.1x,\ x,\ 10x \mid x \in \{0.0030, 0.0035, 0.0040\}\}$ based on the validation performance\footnote{Supposing there are $T$ days of training data, we test the day $T$-1 model on the data of day $T$ to get validation performance.}. 


\subsubsection{Objective Formulation}
The CVR prediction tasks under all attribution mechanisms are formulated as binary classification, where the  loss is defined as:
\begin{equation}
\label{eqn:binary_ce}
\mathcal{L} = - \frac{1}{N} \sum_{i=1}^{N} \Big[ y_i \log p_i + (1 - y_i) \log (1 - p_i) \Big],
\end{equation}
where $N$ is the batch size, $y_i \in \{0, 1\}$ is the ground-truth label of the $i$-th instance, and $p_i \in (0, 1)$ is the predicted conversion probability. 
The overall training objective combines the primary task loss under the target attribution mechanism with auxiliary losses from other attribution mechanisms:
\begin{equation}
\label{eqn:loss} 
\mathcal{L} = \mathcal{L}_{\text{primary}} + \lambda \sum_{k} \mathcal{L}_{\text{aux}}^{(k)},
\end{equation}
where $\mathcal{L}_{\text{primary}}$ denotes the primary task loss, and $\mathcal{L}_{\text{aux}}^{(k)}$ denotes the loss for the $k$-th auxiliary task. 
Each loss term is instantiated as the binary cross-entropy loss in Eq.~\eqref{eqn:binary_ce}.
To find the best $\lambda$ value for each model,  we search over the set $\{0.1, 0.2, 0.3, 0.4\}$ based on validation performance.
We also test an automatic method GCS~\cite{du2020adapting} dynamically setting different weights for each task, but we find it does not make significant contributions, which will be shown in \S~\ref{subsec:atl}.~\looseness=-1

\subsubsection{The Choice of Auxiliary Tasks}
For each target attribution setting, we further search over the combination of auxiliary attribution objectives and report the results under the best setting.
We give the details of our greedy forward-selection search in Appendix~\ref{app:aux}.

\subsection{Main Results and Key Findings}

We present the main results in Table~\ref{tab:main_results} and provide an in-depth analysis by answering the following research questions (RQs):
\begin{itemize}[leftmargin=*] 
\item \textbf{RQ1} (\S\ref{subsubsec:generality}): Does MAL bring substantial performance gains across different target attribution mechanisms?
\item \textbf{RQ2} (\S\ref{subsubsec:user_group}): Which user group gains the most from MAL?
\item \textbf{RQ3} (\S\ref{subsubsec:obj_choice}): Does adding more auxiliary targets from different attribution mechanisms always yield performance growth?
\item \textbf{RQ4} (\S\ref{subsubsec:moae}): Does our MoAE model, which adheres to the key design principles, achieve superior performance?
\end{itemize}

\subsubsection{The Generality of MAL} \label{subsubsec:generality}
Considering that \citet{chen2025see} only demonstrate the contribution of MAL under last-click and DDA attribution mechanisms, it remains unknown whether MAL brings stable performance gains across different target attribution mechanisms.
It is an important question as other attribution mechanisms, e.g., linear attribution, are also widely adopted in recommendation systems.
As shown in Table~\ref{tab:main_results}, the cutting-edge MAL models, e.g., NATAL~\cite{chen2025see} and our \oursys, achieve substantial improvements under all four target attribution settings: last-click, first-click, DDA, and linear.
The performance lifts across diverse attribution mechanisms demonstrate that \textbf{MAL is a general, robust, and highly promising learning paradigm for conversion modeling}.

Furthermore, regarding the improvements under different target attribution mechanisms, we observe that \textbf{last-click and DDA settings see the most significant GAUC} increases at 2.12pt and 1.74pt, respectively.
In contrast, in the \textbf{linear attribution} setting where MAL does not increase the number of positive samples, we observe \textbf{a smaller GAUC uplift} at 0.80pt, which suggests that the causal relationships in DDA labels and the positional information underlying last-click and first-click labels also make contributions.

Notably, in the \textbf{first-click} setting, \textbf{MAL yields the smallest GAUC improvement} at 0.34pt.
We attribute this more modest improvement to the inherently noisy nature of first-click labels.
Specifically, the extended time gap between the initial click and the final conversion introduces many confounding factors (e.g., subsequent promotions) unknown to the model. 
This weakens the label's correlation with the user's original intent, thereby constraining the potential improvement from MAL under the first-click setting.
The fact that the first-click setting yields the lowest absolute AUC and GAUC scores, as shown in Table~\ref{tab:main_results}, underscores the noisy property of the first-click labels, a claim we will further substantiate with empirical evidence in Appendix~\ref{app:first}.

\begin{table}[t]
\centering
\caption{The CVR ranking performance when labels from different attribution methods are leveraged as auxiliary targets. ``None'' denotes the \oursys variant trained without any auxiliary target}
\label{tab:ablation_aux}
\begin{tabular}{@{}l|cc@{}}
\toprule
\textbf{Auxiliary Targets}  & \textbf{AUC} & \textbf{GAUC} \\ \midrule
\rowcolor{gray!20}
\multicolumn{3}{c}{\textbf{Last-Click as Target}} \\ \midrule
None                                &  0.8521  &  0.7429  \\
First-Click                         &  0.8621  &  0.7572  \\
DDA                                 &  0.8543  &  0.7506  \\
Linear                              &  0.8642  &  0.7583  \\
First-Click \& Linear \& DDA        &  \underline{0.8666}  &  \underline{0.7612}  \\ 
First-Click \& Linear \& DDA \& CAT &  \textbf{0.8694}  &  \textbf{0.7636}  \\ 
\midrule
\rowcolor{gray!20}
\multicolumn{3}{c}{\textbf{First-Click as Target}} \\ \midrule
None                                &  \underline{0.8314}  &  \underline{0.6971}  \\
Last-Click                          &  \textbf{0.8325}  &  \textbf{0.7003}  \\
DDA                                 &  0.8248  &  0.6932  \\
Linear                              &  0.8182  &  0.6886  \\
Last-Click \& DDA \& Linear         &  0.8171  &  0.6898  \\ 
Last-Click \& DDA \& Linear  \& CAT &  0.8182  &  0.6878  \\ 
\midrule
\rowcolor{gray!20}
\multicolumn{3}{c}{\textbf{DDA as Target}} \\ \midrule
None                                       &  0.8447  &  0.7481  \\
Last-Click                                 &  0.8467  &  0.7494  \\
First-Click                                &  0.8502  &  0.7530  \\
Linear                                     &  0.8520  &  0.7570  \\
Last-Click \& First-Click \& Linear        &  \underline{0.8521}  &  \underline{0.7592}  \\ 
Last-Click \& First-Click \& Linear \& CAT &  \textbf{0.8576}  &  \textbf{0.7644}  \\ 
\midrule
\rowcolor{gray!20}
\multicolumn{3}{c}{\textbf{Linear as Target}} \\ \midrule
None                                     &  0.8524  &  0.7611  \\
Last-Click                               &  0.8545  &  0.7649  \\
First-Click                              &  0.8534  &  0.7618  \\
DDA                                      &  0.8535  &  0.7632  \\
Last-Click \& First-Click \& DDA         &  \underline{0.8552}  &  \underline{0.7661}  \\ 
Last-Click \& First-Click \& DDA  \& CAT &  \textbf{0.8568}  &  \textbf{0.7682}  \\ 
\bottomrule
\end{tabular}
\end{table}

\subsubsection{Impact Analysis on User Groups}
\label{subsubsec:user_group}

To answer \textbf{RQ2}, we present the AUC lift brought by our \oursys over the single-attribution baseline across \textbf{user groups categorized by conversion path complexity} in Figure~\ref{fig:user_groups}.
Specifically, the conversion path complexity is proxied by the ratio of the number of positive samples under the linear-click setting to that under the last-click setting. 
A higher value of this ratio indicates that users click more ads on average before conversion, signifying a longer conversion path.

We observe that \textbf{multi-attribution learning delivers greater AUC improvements for users with more complex conversion paths}, a trend that holds across most of the evaluated target attribution mechanisms: last-click, linear, and DDA.
This correlation suggests that \textbf{MAL is particularly beneficial for users with complex multi-touch conversion journeys}, 
where richer information gains underlie multi-attribution labels.
In contrast, when first-click is the target attribution mechanism, the AUC improvements brought by MAL do not exhibit a clear monotonic trend with respect to conversion path complexity,  which is likely caused by the inherently noisy nature of first-click labels.

\subsubsection{The Effect of Auxiliary Objective Choice}
\label{subsubsec:obj_choice}

We progressively add different auxiliary targets to \oursys under each target attribution mechanism and report the resulting model performance in Table~\ref{tab:ablation_aux}, where we make two key observations as follows.

\textbf{First, the model's performance lift scales with the number of auxiliary targets} under most of target attribution mechanisms, and the Cartesian product auxiliary target (CAT)~\cite{chen2025see} yields a further significant improvement, demonstrating \textit{the potential of incorporating more sophisticated auxiliary targets}.

\textbf{Second, in contrast, the first-click CVR prediction model is sensitive to the choice of auxiliary objectives}.
Concretely, only the last-click auxiliary target yields significant performance gains, while other choices of auxiliary targets undermine model performance.
We speculate that this phenomenon is due to the unique position of the last-click target.
Because it is temporally closest to the conversion, it offers the greatest discrepancy from the first-click signal, which in turn helps the model learn the user's intrinsic intent more effectively.

\subsubsection{The Superiority of Our \oursys Model}
\label{subsubsec:moae}

As shown in Table~\ref{tab:main_results}, \textbf{\oursys consistently outperforms all baselines under all four target attribution mechanisms.}
Compared with the existing state-of-the-art model NATAL~\cite{chen2025see}, \oursys achieves GAUC gains of +0.23pt, +0.13pt, +0.27pt, and +0.39pt under last-click, first-click, DDA, and linear settings, respectively.
We further validate the GAUC improvement with five random seeds under the last-click setting, as reported in Appendix~\ref{app:multi_seed}.
The superiority of \oursys substantiates the power of combining the two architecture design principles in \S~\ref{subsubsec:moae}, namely \textbf{(1) fully learning multi-attribution knowledge} underlying multi-attribution targets and \textbf{(2) fully leveraging multi-attribution knowledge} to maximize the main task performance.
We are confident that these findings will serve as valuable guidelines for future MAL architectural designs.

\section{Further Discussion}

In this section, we discuss two research questions that are not covered by the major experiments:
\begin{itemize}[leftmargin=*] 
\item \textbf{RQ5}: Does the performance growth brought by MAL come from the increase of model parameters or the information gain underlying multi-attribution labels?
\item \textbf{RQ6}: How do auxiliary-task learning (ATL) techniques affect the performance of MAL models?
\end{itemize}

\subsection{Effect of Parameter Growth}

Given that MAL models such as NATAL~\cite{chen2025see} and our \oursys introduce more parameters for fitting multiple attribution labels compared to the single-attribution baseline, the performance growth of MAL models may be attributed to \textbf{parameter scaling} rather than  \textbf{multi-attribution knowledge utilization}.

\begin{table}[!t]
\centering
\caption{The CVR ranking performance with/without auxiliary labels from multiple attribution mechanisms (mal). Last-click is the target attribution mechanism.}
\label{tab:ablation_param}
\begin{tabular}{@{}l|cc@{}}
\toprule
\textbf{Method/Metrics} & \textbf{AUC $\uparrow$}   & \textbf{GAUC $\uparrow$} \\ \midrule
Base & 0.8494 & 0.7424 \\ \midrule
Shared-Bottom & 0.8628 \textcolor{Green}{(+1.34pt)} & 0.7571 \textcolor{Green}{(+1.47pt)} \\
{\ \ \bf -} w/o mal & 0.8497 \textcolor{gray}{(+0.03pt)} & 0.7420 \textcolor{gray}{(-0.04pt)}\\ \midrule
MMoE & 0.8642 \textcolor{Green}{(+1.48pt)} & 0.7571 \textcolor{Green}{(+1.47pt)} \\
{\ \ \bf -} w/o mal & 0.8501 \textcolor{gray}{(+0.07pt)} & 0.7428 \textcolor{gray}{(+0.04pt)}\\ \midrule
PLE  & 0.8645 \textcolor{Green}{(+1.51pt)} & 0.7583 \textcolor{Green}{(+1.59pt)} \\
{\ \ \bf -} w/o mal & 0.8494 \textcolor{gray}{(0.00pt)} & 0.7424 \textcolor{gray}{(0.00pt)}\\ \midrule
HoME & 0.8661 \textcolor{Green}{(+1.67pt)} & 0.7594 \textcolor{Green}{(+1.70pt)} \\
{\ \ \bf -} w/o mal & 0.8523 \textcolor{gray}{(+0.29pt)} & 0.7421 \textcolor{gray}{(-0.03pt)}\\ \midrule
NATAL & 0.8671 \textcolor{Green}{(+1.77pt)} & 0.7613 \textcolor{Green}{(+1.89pt)} \\ 
{\ \ \bf -} w/o mal & 0.8527 \textcolor{gray}{(+0.33pt)} & 0.7437 \textcolor{gray}{(+0.13pt)}\\ \midrule
\oursys \textbf{(Ours)} & 0.8694 \textcolor{Green}{(+2.00pt)} & 0.7636 \textcolor{Green}{(+2.12pt)} \\
{\ \ \bf -} w/o mal & 0.8521 \textcolor{gray}{(+0.27pt)} & 0.7429 \textcolor{gray}{(+0.05pt)}\\ 
\bottomrule
\end{tabular}
\end{table}

To verify the source of performance gains brought by MAL, we disentangle the effect of multi-attribution labels (\textbf{mal} for short) from parameter scaling by setting the auxiliary tasks weights to zeros, and list the resulting model metrics in Table~\ref{tab:ablation_param}.

We observe that \textbf{vanilla parameter scaling yields no or marginal GAUC improvements}, and \textbf{the performance growth brought by MAL mainly results from multi-attribution supervision signals.}
Specifically, when the auxiliary task weights are set to zero, namely the ``w/o mal'' variants in Table~\ref{tab:ablation_param}, although the parameter scales of resulting models are larger than the single-attribution Base and equal to corresponding MAL models, they significantly underperform the MAL counterparts learning from auxiliary labels, and perform almost on par with the Base model.
The trend is consistent across the multi-task family (Shared-Bottom, MMoE, PLE, and HoME), NATAL, and our \oursys model, suggesting that harnessing multi-attribution knowledge underlying diverse attribution mechanisms contributes to the performance advantage of MAL models over the single-attribution baseline.

\subsection{Effect of Auxiliary-Task Learning Methods} \label{subsec:atl}

Auxiliary-task learning (ATL)~\cite{du2020adapting,liuauto,yu2020gradient} aims to maximize the target task performance by leveraging the knowledge underlying auxiliary tasks, which aligns with our MAL problem setup.
ATL is a developed research area with various approaches, such as gradient manipulation~\cite{yu2020gradient,wanggradient}, auxiliary loss weight adjustment~\cite{lin2019adaptive,du2020adapting,shi2020auxiliary}, and auxiliary task grouping~\cite{fifty2021efficiently,song2022efficient}, and we believe that incorporating ATL techniques is promising for future work on multi-attribution learning.
In this study, we take the first step to investigate the effect of ATL techniques on multi-attribution learning models.
Concretely, we implement the classic  GCS~\cite{du2020adapting} and PCGrad~\cite{yu2020gradient} approaches as \textbf{optional training features in our PyMAL} library, and examine their effects on MAL models.
GCS adjusts auxiliary loss weights using gradient similarities between target and auxiliary targets, and PCGrad eliminates gradient conflicts through gradient projection.

We present the results in Table~\ref{tab:ablation_grad} and make two key observations.
\textbf{First}, GCS and PCGrad yield small performance gains when attached to weak baselines such as Shared-Bottom and MMoE, showing their effectiveness in mitigating negative transfer, namely the negative effects of auxiliary tasks on the target task.
\textbf{Second}, when applied to strong MAL models such as NATAL and our \oursys tailored for maximizing the target task performance, GCS and PCGrad show marginal negative effects.
Overall, our \oursys without GCS or PCGrad achieves the best performance among all model variants.
These results indicate that \textbf{well-designed architectures are more effective than ATL techniques based on gradient projection and loss weight adjustment in multi-attribution learning for CVR prediction}.
We believe that it is easy to incorporate more advanced ATL approaches to our \textbf{PyMAL} library, which are promising for future study.

\begin{table}[!t]
\centering
\caption{The effect of ATL techniques GCS and PCGrad.}
\label{tab:ablation_grad}
\begin{tabular}{@{}l|cc@{}}
\toprule
\textbf{Method/Metrics} & \textbf{AUC $\uparrow$}   & \textbf{GAUC $\uparrow$} \\ \midrule
Shared-Bottom & 0.8628 & 0.7571 \\
{\ \ \bf +} GCS & 0.8632 \textcolor{Green}{(+0.04pt)} & 0.7589 \textcolor{Green}{(+0.18pt)}\\ 
{\ \ \bf +} PCGrad & 0.8636 \textcolor{Green}{(+0.08pt)} & 0.7591 \textcolor{Green}{(+0.20pt)}\\
\midrule
MMoE & 0.8642 & 0.7571 \\ 
{\ \ \bf +} GCS & 0.8636 \textcolor{Red}{(-0.06pt)} & 0.7590 \textcolor{Green}{(+0.19pt)}\\ 
{\ \ \bf +} PCGrad & 0.8640 \textcolor{Red}{(-0.02pt)} & 0.7573 \textcolor{Green}{(+0.02pt)}\\
\midrule
PLE & 0.8645 & 0.7583 \\
{\ \ \bf +} GCS & 0.8638 \textcolor{Red}{(-0.07pt)} & 0.7572 \textcolor{Red}{(-0.11pt)}\\ 
{\ \ \bf +} PCGrad & 0.8636 \textcolor{Red}{(-0.09pt)} & 0.7568 \textcolor{Red}{(-0.15pt)}\\
\midrule
HoME & 0.8661 & 0.7594 \\
{\ \ \bf +} GCS & 0.8626 \textcolor{Red}{(-0.35pt)} & 0.7569 \textcolor{Red}{(-0.25pt)}\\ 
{\ \ \bf +} PCGrad & 0.8653 \textcolor{Red}{(-0.08pt)} & 0.7609 \textcolor{Green}{(+0.15pt)}\\
\midrule
NATAL & 0.8671 & 0.7613 \\
{\ \ \bf +} GCS & 0.8628 \textcolor{Red}{(-0.43pt)} & 0.7571 \textcolor{Red}{(-0.42pt)}\\ 
{\ \ \bf +} PCGrad & 0.8663 \textcolor{Red}{(-0.08pt)} & 0.7618 \textcolor{Green}{(+0.05pt)}\\
\midrule
\oursys \textbf{(Ours)} & 0.8694 & 0.7636 \\
{\ \ \bf +} GCS & 0.8673 \textcolor{Red}{(-0.21pt)} & 0.7604 \textcolor{Red}{(-0.32pt)}\\
{\ \ \bf +} PCGrad & 0.8680 \textcolor{Red}{(-0.14pt)} & 0.7625 \textcolor{Red}{(-0.11pt)}\\
\bottomrule
\end{tabular}
\end{table}

\section{Conclusion}

In this work, we introduce \data, the first public CVR prediction benchmark providing \textbf{conversion labels derived from multiple attribution mechanisms}. 
This dataset enables systematic investigation into \textbf{multi-attribution learning (MAL)}, a promising paradigm previously hindered by the unavailability of multi-attribution conversion labels.
Through comprehensive experiments on \data using our open-source \textbf{PyMAL} library, we gather three key insights. 
\textbf{First}, MAL consistently enhances model performance across diverse target attribution settings, with the most significant gains observed for users exhibiting long and complex conversion paths.
\textbf{Second}, the choice of auxiliary objectives in MAL is crucial and task-dependent.
\textbf{Third}, effective MAL architectures must satisfy two core design principles: (i) fully learn the knowledge embedded in multi-attribution labels, and (ii) effectively leverage this knowledge in a main-task-prioritized manner.
Motivated by these insights, we propose \oursys, an effective MAL model that integrates an MoE backbone for knowledge acquisition with an asymmetric transfer module for main-task-centric utilization, whose superiority is demonstrated by extensive empirical evaluation.

We believe the \data benchmark, the \textbf{PyMAL} toolkit, and the gathered actionable insights will serve as valuable resources for the research community, accelerating future innovation in multi-attribution learning and more holistic conversion modeling.
For future work, we believe that incorporating stronger foundation model capabilities, such as chain-of-thought reasoning~\cite{wei2022chain}, and more comprehensive conversion objectives~\cite{zeng2025clickabuyb} holds great promise.

\begin{acks}
This work is funded by Fundamental and Interdisciplinary Disciplines Breakthrough Plan of the Ministry of Education of China (JYB2025XDXM902), National Natural Science Foundation of China (Grant No. 62506158 and No. 62441234), Basic Research Program of Jiangsu (BK20251183), and CCF-ALIMAMA TECH Kangaroo Fund (No. CCF-ALIMAMA of 2025004).
\end{acks}

\bibliographystyle{ACM-Reference-Format}
\balance
\bibliography{references}

\appendix

\section{Details of Auxiliary Objective Search}
\label{app:aux}
For each model under each attribution setting, we start from the primary objective alone and greedily add auxiliary objectives in a forward-selection manner:  (1) at each step, we temporarily add one candidate auxiliary objective, train the model, and measure the GAUC improvement on the validation day;  (2) we then select the auxiliary objective that brings the largest positive gain and permanently include it into the auxiliary objective set;  (3) this process is repeated on the remaining candidates until either all auxiliary objectives have been used or no further GAUC improvement is observed. This greedy search is conducted independently under each target attribution mechanism. 

In our experiments, this greedy search yields consistent patterns across settings. For all models under the last-click, linear, and DDA attribution mechanisms, the optimal configuration includes all candidate auxiliary objectives. Under the first-click attribution mechanism, however, the best configuration for all models uses only the last-click objective as an auxiliary signal.

\section{Analysis of First-Click Labels}
\label{app:first}
We provide additional evidence of the noisier nature of first-click labels by examining their inconsistency with data-driven attribution (DDA) and linear attribution, summarized in Table~\ref{tab:mta_minus_linear}.
The table reports the mean difference between the DDA weight and the linear weight for positive labels (MML). 
Under the last-click setting, MML is positive ($+0.05$), indicating that last-click positive samples tend to receive more weight under DDA than under the simpler linear rule—i.e., the DDA view broadly agrees that these touchpoints are important. 
In contrast, under the first-click setting, MML becomes negative ($-0.03$), meaning that clicks labeled as positive by first-click are, on average, assigned less weight by DDA than by linear attribution. Since DDA is learned from causal signals along the entire path, this finding suggests that many first-click ``positives'' are not strongly supported as influential touchpoints and are even down-weighted relative to a naive linear split. 
This comparison indicates that first-click labels are noisier and harder to predict.

\begin{table}[thp] 
\centering
\caption{The MML metric comparison between the first-click and last-click positive samples.}
\label{tab:mta_minus_linear}
\begin{tabular}{@{}l|cc@{}}
\toprule
\textbf{} &
\textbf{Last-Click} &
\textbf{First-Click}\\
\midrule
\textbf{MML}  & 0.05 & -0.03 \\
\bottomrule
\end{tabular}
\end{table}

\section{Multi-Seed Validation}
\label{app:multi_seed}
To further examine the stability of the performance improvement, we repeat the representative last-click experiment with five random seeds and compare \oursys with the strongest existing MAL baseline \presys.
As shown in Table~\ref{tab:multi_seed}, \oursys consistently outperforms \presys across runs, achieving an average GAUC gain of $+0.18$pt.
A significance test on GAUC gives $p \approx 0.006$, suggesting that the improvement is statistically reliable under this setting.

\begin{table}[h]
\centering
\caption{Results on the last-click setting over five random seeds.}
\label{tab:multi_seed}
\begin{tabular}{lcc}
\toprule
Model & AUC & GAUC \\
\midrule
\presys & $0.8673 \pm 0.0002$ & $0.7612 \pm 0.0008$ \\
\oursys & $0.8691 \pm 0.0005$ & $0.7630 \pm 0.0002$ \\
\midrule
$\Delta$ & $+0.18$pt & $+0.18$pt \\
\bottomrule
\end{tabular}
\end{table}

\section{Details of DDA Label Generation}
\label{app:dda}
The data-driven attribution (DDA) labels in \data are generated by CausalMTA~\cite{yao2022causalmta}, a causal multi-touch attribution model that estimates conversion credits from user conversion paths.
Unlike rule-based mechanisms such as last-click, first-click, and linear attribution, CausalMTA estimates the counterfactual contribution of each touchpoint to the final conversion.

CausalMTA first learns a debiased CVR predictor for counterfactual conversion estimation, which is required by Shapley-value-based attribution.
To mitigate the confounding effect of user preference on both ad exposure and conversion behavior, CausalMTA combines two debiasing modules: a journey reweighting module based on a variational recurrent autoencoder for static user attributes, and a causal recurrent predictor with a gradient reversal layer for dynamic behavior features.
The learned debiased predictor is then used to estimate the Shapley-value attribution credit of each touchpoint, where the DDA weight reflects the marginal contribution of the corresponding click to the conversion probability.

In our benchmark construction, we apply CausalMTA to conversion paths within the fixed attribution window and use the resulting attribution credits as the DDA labels.
The same CausalMTA-based attribution system has served large-scale online traffic in Alibaba's advertising platform for several years, supporting the reliability of the generated DDA labels.

\section{Details of CAT Label Construction}
\label{app:cat}
The Cartesian-style Auxiliary Target (CAT)~\cite{chen2025see} transforms the set of $N$ binary attribution labels into a single multi-class label by enumerating all possible joint outcomes across attribution mechanisms.

Concretely, for each sample we start from the binary labels
\[
\mathbf{A} = \big(A_0, A_1, \dots, A_{N-1}\big), \quad A_i \in \{0,1\},
\]
where $A_i$ is the conversion indicator under the $i$-th attribution mechanism (e.g., last-click, first-click, DDA, linear). The CAT label is then defined as the integer encoding of $\mathbf{A}$ in little-endian binary form:
\begin{equation}
\mathbf{O} = \sum_{i=0}^{N-1} A_i \cdot 2^i, \quad \mathbf{O} \in \{0,1,\dots,2^N-1\}.
\label{eq:cat_def}
\end{equation}
Algorithm~\ref{alg:cat_app} gives the corresponding pseudo-code. In our experiments with four attribution mechanisms (last-click, first-click, DDA, linear), CAT becomes a 16-class classification problem. For instance, the combination $(1,0,1,0)$ is mapped to class
\[
\mathbf{O} = 1 \cdot 2^0 + 0 \cdot 2^1 + 1 \cdot 2^2 + 0 \cdot 2^3 = 5.
\]

\begin{algorithm}
\caption{Computation of the \textbf{CAT} Label}
\label{alg:cat_app}
\begin{algorithmic}[1]
\REQUIRE Array $\mathbf{A}$ of length $N$ with elements in $\{0, 1\}$, i.e., the conversion indicators under $N$ attribution mechanisms
\ENSURE Integer CAT label $\mathbf{O} \in [0, 2^N - 1]$
\STATE Initialize $\mathbf{O} \gets 0$
\FOR{$i = 0$ to $N-1$}
    \STATE $\mathbf{O} \gets \mathbf{O} + \mathbf{A}[i] \cdot 2^i$
\ENDFOR
\end{algorithmic}
\end{algorithm}

During training, we treat $\mathbf{O}$ as the target of a $(2^N)$-way classification task, optimized with a standard softmax cross-entropy loss. The corresponding conversion knowledge vector $\mathcal{K}_{\text{CAT}}$ is extracted from the CAT prediction tower and fused into the overall conversion representation $\mathcal{K}$ learned by AKA. At inference time, we do not use the CAT logits directly; only the enriched representation $\mathcal{K}$ is consumed by the primary target predictor.

\end{document}